\documentclass[letterpaper]{article} 
\usepackage{aaai24}  
\usepackage{times}  
\usepackage{helvet}  
\usepackage{courier}  
\usepackage[hyphens]{url}  
\usepackage{graphicx} 
\usepackage{natbib}  
\usepackage{caption} 
\usepackage{fancyhdr}
\usepackage{cite}
\usepackage{amsmath,amssymb,amsfonts}
\usepackage{algorithmic}
\usepackage{graphicx}
\usepackage{textcomp}
\usepackage{xcolor}
\usepackage{booktabs}
\usepackage{multirow}
\usepackage{makecell}

\usepackage{microtype}
\usepackage{graphicx}
\usepackage{subfigure}
\usepackage{booktabs} 
\usepackage[nameinlink]{cleveref}

\usepackage{enumitem}
\newlist{propenum}{enumerate}{1} 
\setlist[propenum]{label=\alph*{\rm)}, ref=\theproposition(\alph*)}
\crefalias{propenumi}{proposition}
\newlist{corenum}{enumerate}{1} 
\setlist[corenum]{label=\alph*{\rm)}, ref=\thecorollary(\alph*)}
\crefalias{corenumi}{corollary}
\newlist{lemenum}{enumerate}{1} 
\setlist[lemenum]{label=\alph*{\rm)}, ref=\thelemma(\alph*)}
\crefalias{lemenumi}{lemma}

\usepackage{amsthm,thmtools}

\declaretheorem[name=Example,style=definition,qed=\qedsymbol]{example}

\usepackage{ragged2e}

\usepackage{amsthm,thmtools}

\usepackage[nameinlink]{cleveref}

\usepackage{times}
\usepackage{soul}
\usepackage{url}
\usepackage[utf8]{inputenc}
\usepackage{caption}
\usepackage{graphicx}
\usepackage{amsmath}
\usepackage{amsthm}
\usepackage{booktabs}
\usepackage{algorithm}
\usepackage{algorithmic}
\usepackage[switch]{lineno}

\usepackage{booktabs}
\usepackage{multirow}
\usepackage{bbding}

\title{Combinatorial Optimization with Automated Graph Neural Networks}

\author{
    Yang Liu\textsuperscript{\rm 1}, 
    Peng Zhang\textsuperscript{\rm 4}, 
    Yang Gao\textsuperscript{\rm 3}, 
    Chuan Zhou\textsuperscript{\rm 1},
    Zhao Li\textsuperscript{\rm 2}, 
    Hongyang Chen\textsuperscript{\rm 5}
}
\affiliations{

    \textsuperscript{\rm 1}Academy of Mathematics and Systems Science, Chinese Academy of Sciences\\
    \textsuperscript{\rm 2}Hangzhou Yugu Technology Co., Ltd
    \textsuperscript{\rm 3}Zhejiang University
    \textsuperscript{\rm 4}Guangzhou University
    \textsuperscript{\rm 5}Zhejiang Lab\\
    \{liuyang2020, zhouchuan\}@amss.ac.cn
}

\usepackage{bibentry}

\begin{document}

\maketitle

\begin{abstract}
    In recent years, graph neural networks (GNNs) have become increasingly popular for solving NP-hard combinatorial optimization (CO) problems, such as maximum cut and maximum independent set. The core idea behind these methods is to represent a CO problem as a graph and then use GNNs to learn the node/graph embedding with combinatorial information. Although these methods have achieved promising results, given a specific CO problem, the design of GNN architectures still requires heavy manual work with domain knowledge. Existing automated GNNs are mostly focused on traditional graph learning problems, which is inapplicable to solving NP-hard CO problems. To this end, we present a new class of \textbf{AUTO}mated \textbf{G}NNs for solving \textbf{NP}-hard problems, namely \textbf{AutoGNP}. We represent CO problems by GNNs and focus on two specific problems, i.e., mixed integer linear programming and quadratic unconstrained binary optimization. The idea of AutoGNP is to use graph neural architecture search algorithms to automatically find the best GNNs for a given NP-hard combinatorial optimization problem. Compared with existing graph neural architecture search algorithms, AutoGNP utilizes two-hop operators in the architecture search space. Moreover, AutoGNP utilizes simulated annealing and a strict early stopping policy to avoid local optimal solutions. Empirical results on benchmark combinatorial problems demonstrate the superiority of our proposed model.
\end{abstract}

\section{Introduction}

Combinatorial optimization (CO) refers to a class of discrete optimization that searches for the minimum of an objective function on a discrete variable set. CO plays an important role in a diverse range of applications, such as transportation \cite{wang2021deep} and scheduling \cite{verma2021comprehensive}. Most CO problems are NP-hard and thus computationally challenging. Due to the high computational complexity of these problems, exact solutions are anticipated to take exponential amounts of time. Heuristic methods are frequently used to find satisfactory sub-optimal solutions in a reasonable amount of computing time, without a provable optimal guarantee of the solutions attained. While there emerge a large number of new algorithms for various CO problems, there is a tendency to utilize machine learning methods to solve CO problems \cite{cappart2023combinatorial}, such as utilizing graph neural networks (GNNs) as the solution.

Recently, GNNs have been used to solve CO problems and have shown promising results on different kinds of graph-structured optimization problems (e.g., mixed integer linear programming). GNNs are suitable for representing CO problems due to their merit of permutation invariance, which means the CO problems and the solutions are not fundamentally altered by the operator of permutations applied to the variables. There exist several works \cite{Glover2022QuantumBA,gasse2019exact} to show promise for solving combinatorial optimization in a differentiable, end-to-end framework.

Although existing GNN methods have shown more promising results in solving CO problems than previous methods such as local search and simulated annealing, the architecture of GNNs still remains hand-crafted and limits the capability for the tasks. In conventional GNNs such as GCN, GAT, and GraphSAGE, the selection of these components is based on expert knowledge and implicit guidelines. Thus, optimizing combinations of these components is challenging. Building the most effective GNN models for solving a combinatorial task requires evaluating a large number of GNN architectures and then selecting the best one. For NP-hard CO problems, it is urgent to design accurate GNNs with minimal human manual effort, considering the computations of CO problems are also time-consuming. An intuitive idea is to use an automated graph neural network such as GraphNAS~\cite{Gao2019GraphNASGN} to replace the hand-crafted GNN architectures. In fact, there has been a line of works that extend GraphNAS to solve graph learning problems in the past few years \cite{Gao2019GraphNASGN,zhou2022auto,lai2020policy,huan2021search}, and empirical results have shown that automated graph learning is capable of finding the best GNN architectures based on a given validation graph dataset. However, the above methods have not touched the problem of solving CO problems using automated GNNs, which is our main focus.

In this paper, we attempt to answer the following question: \emph{``Given a combinatorial optimization problem which is NP-hard, how to design the best graph neural architecture to represent the problem, to improve the performance on downstream tasks?"}. To design a graph neural architecture search model for solving CO problems, we propose a new automated graph neural network \textbf{AutoGNP}  as the solution.  Different from the original automated GNNs such as GraphNAS for graph learning tasks, GNNs for CO problems can be taken as unsupervised learning in a discrete feature space. In terms of search algorithms, we expect to design an efficient differentiable learning search algorithm, because the search space of GNNs is generally very large and the training of the searching algorithm converges slowly. Furthermore, we represent CO problems by GNNs, especially for the mixed integer linear programming problem and the quadratic unconstrained binary optimization problem. We build the search space of AutoGNP by including the functions (operators) used in existing popular GNNs and add a two-hop message passing operator for AutoGNP. Also, we design two training methods, i.e., simulated annealing and strict early stopping policy,  to avoid failing to locally optimal solutions. We build an end-to-end learning framework based on AutoGNP to solve NP-hard CO problems and evaluate the performance of benchmark datasets by comparing them with existing competitive baselines. Our contributions are summarized as follows:
\begin{itemize}
\item {\bf Automated Framework: AutoGNP.}\; We propose AutoGNP, an automated graph neural network model designed to address COPs. Furthermore, we present a novel differentiable gradient algorithm tailored to identifying optimal graph neural architectures specifically for COPs.

\item {\bf Representation for COP instance.}\; We consider it to represent two specific COPs, i.e., mixed integer linear programming and quadratic unconstrained binary optimization. We show that it is suitable for handling these two large-scale COPs.

\item  {\bf Experimental evaluation.}\; Empirical evaluations of AutoGNP on various benchmark tasks and datasets over combinatorial domains demonstrate the superiority of our method, which improves the performance of existing GNN architectures in solving NP-hard CO problems.
\end{itemize}

\section{Related works}\label{appendix: related}

In this section, we introduce closely related works, including graph neural networks (GNNs), GNN-based models for solving combinatorial optimization, and automated graph neural architecture search algorithms. 

{\bf GNN for combinatorial optimization.} To solve  NP-hard problems, a number of machine learning models have been developed to improve existing algorithms \cite{Bengio2018MachineLF,liu2021decision}. Graph neural networks (GNNs) serve as a fundamental modular block for combinatorial optimization, either as solvers or by enhancing classical solvers.

Several recent works \cite{Bengio2018MachineLF} have proposed GNN-based approaches for problems such as the Traveling Salesman Problem (TSP) and Vehicle Routing Problem (VRP). For the TSP, \cite{vinyals2015pointer} proposed a Pointer Network that learns to point to nodes in a sequence to produce a tour. \cite{bello2016neural} introduced a Reinforcement Learning approach using a policy gradient method with a GNN as the policy network. \cite{hammouri2018dragonfly} presented a differentiable relaxation of the TSP that can be optimized with gradient-based methods. \cite{nazari2022dadam} introduced an attention-based GNN that iteratively refines candidate tours. For the VRP, \cite{nazari2018reinforcement} proposed a GNN-based model with a differentiable k-hop routing mechanism to produce vehicle routes. 

GNNs have been successfully applied to obtain strong baselines and state-of-the-art results on several combinatorial optimization problems. They show great promise for solving such problems in a differentiable, end-to-end manner. These frameworks can be categorized into reinforcement learning methods and imitation learning methods.  Reinforcement learning (RL) is a popular framework to solve such undifferentiated processes \cite{Gasse2019ExactCO} such as integer programming. RL is independent of labeled training sets, instead, agents in a RL system attempt to explore the environment, such as a graph, by performing actions. On the other hand, Imitation learning (IL) was proposed by \cite{Ross2013InteractiveLF} and it can lead or imitate the ``export" rules in traditional algorithms. In this way, it can provide rules with excellent performance without other professional knowledge. For example, the strong branch rule in Branch and Bound \cite{Morrison2016BranchandboundAA}, a classical exact algorithm to solve integer programming, can be imitated by deep learning models many works called ``Learning to Branch" \cite{Khalil2016LearningTB}. There also emerge other imitation-based works such as ``Learning to dive" \cite{Paulus2023LearningTD}, ``Learning to cut" \cite{Berthold2022LearningTU}, and so on.

{\bf Graph neural architecture search.} Graph neural architecture search, known as GraphNAS\cite{Gao2019GraphNASGN}, represents a seed work for automatically designing the architecture of graph neural networks (GNNs) by searching through different candidate structures \cite{Zhang2021AutomatedML}. In recent years, there has been significant progress in the field of graph neural architecture search, and several approaches have been proposed to improve the performance  \cite{Gao2019GraphNASGN,zhou2022auto,lai2020policy,huan2021search}. Existing works have focused on combining multiple search strategies to improve the overall performance of the search process. For example, several methods use both reinforcement learning \cite{Gao2019GraphNASGN} and evolutionary algorithms \cite{zhou2022auto} to jointly search for the optimal architecture. Overall, GraphNAS has shown promising results in automatically designing accurate graph neural networks. There is still much research effort  \cite{wang2023zeroth} in this area to further improve its efficiency and accuracy.

\begin{table*}[t]
\small
    \caption{ \small{Comparison to the similar framework with several properties. EA and RL are short for evolutionary algorithms and reinforcement learning. Heter, Homo, and Normal denote heterogeneous graphs (with labels), homogeneous graphs (with labels), and ordinary graphs (without labels).} } \label{table: properties}
    \vskip 0.05in
    \centering
         \begin{tabular}{lccllc} 
        \toprule
        \multirow{2}{*}{\textbf{ Framework }}
        & \multicolumn{5}{c}{\textbf{ Properties }} \\
        & Efficient & Differentiable & Input Graph & Search Algorithm & CO problems \\
        \midrule
        GraphNAS \cite{Gao2019GraphNASGN}
        & \XSolidBrush & \XSolidBrush & Homo & RL & \XSolidBrush \\
        Auto-GNN \cite{zhou2022auto}
        & \XSolidBrush & \XSolidBrush & Homo & EA+RL & \XSolidBrush \\
        Policy-GNN \cite{lai2020policy}
        & \XSolidBrush & \XSolidBrush & Homo & RL & \XSolidBrush \\
        \midrule
        SANE \cite{huan2021search}
        & \Checkmark & \Checkmark & Heter & Differentiable & \XSolidBrush \\
        DiffMG \cite{ding2021diffmg} & \Checkmark & \Checkmark & Heter & Differentiable & \XSolidBrush \\
        AutoGEL \cite{Wang2021AutoGELAA} 
        & \Checkmark & \Checkmark & Heter & Differentiable & \XSolidBrush \\
        GraphGym \cite{you2020design} 
        & \Checkmark & \Checkmark & Homo & Random & \XSolidBrush \\
        ZARTS \cite{wang2022zarts} 
        & \Checkmark & \Checkmark & Homo & Zeroth-order & \XSolidBrush \\
        NAC \cite{Xu2023DoNT} & \Checkmark & \Checkmark & Homo & Differentiable & \XSolidBrush \\
        \midrule
        AutoGNP (ours) & \Checkmark & \Checkmark & Normal & Differentiable & \Checkmark \\
        \bottomrule
         \end{tabular}
    \end{table*}

\section{Preliminaries and problem setup}

{\bf Notations.} We denote a graph as $\mathcal{G}$ and  its edges and nodes as $\mathcal{E}$ and $\mathcal{V}$ respectively. 
We represent an ordinary graph as a set of edges $\mathcal{E} = \{ (v_i, v_j) \mid v_i, v_j \in \mathcal{V} \}$, where $n$ is the number of observed edges. For each node $v$ and edge $e$, we use their bold version $\mathbf{v}$ and $\mathbf{e}$ to denote their embeddings. We use bold capital letters. e.g., $\mathbf{A}, \mathbf{B}, \mathbf{W}, \mathbf{Q}$ to denote matrices and use $\| \cdot \|$ to denote the Euclidean norm of vectors or the Frobenius norm of matrices. Due to space limit, we summarize the main symbols and notations used in this paper in \Cref{table: notation}.

\subsection{Problem setup}\label{section: setup}

In this paper, we consider two types of NP-hard combinatorial optimization problems, i.e., mixed integer linear programming (MILP) and quadratic unconstrained binary optimization (QUBO).

\paragraph{MILP.} A general MILP problem can be defined as follows, 
\begin{align}\label{equation: MILP}
    \mathbf{x}^* = \underset{ \mathbf{x} \in \mathbb{Z}^p \times \mathbb{R}^{n-p} }{\arg \min} \, \left\{ \mathbf{c}^T \mathbf{x} \, | \,  \mathbf{A}\mathbf{x} \leq \mathbf{b}, \mathbf{l} \leq \mathbf{x} \leq \mathbf{u} \right\}, 
\end{align}
where $\mathbf{c} \in \mathbb{R}^n$, $\mathbf{A} \in \mathbb{R}^{m \times n}$ are constraint coefficient matrices and $\mathbf{b} \in \mathbb{R}^{m}$ is a constraint right-hand-side vector,  $\mathbf{l}, \mathbf{u} \in \mathbb{R}^{n}$ are the lower and upper bounds for variables, respectively.

\begin{example}[Stock market portfolio optimization]
    Portfolio optimization is a classical quadratic programming problem which aims to choose $k$ stocks out of $N$  with learnable parameters that maximize the penalized objective function as follows:
    \begin{align}
        \max_{\mathbf{x}} \quad & \mathbf{p}^{{T}} \mathbf{x} \\
        \text{s.t.} \quad & \mathbf{x}^{{T}} \mathbf{Q} \mathbf{x} \leq M. 
    \end{align}
    In this problem, vector $\mathbf{p}$ and matrix $\mathbf{Q}$ are unknown and we shall learn from the data.
\end{example}

\paragraph{QUBO.} A QUBO model can be described as minimizing a QUBO Hamiltonian $H$ as follows:
\begin{align}\label{equation: QUBO}
    \mathbf{x}^* = \underset{ \mathbf{x} \in \{0,1\}^n }{\arg \min} \, H(\mathbf{x};\mathbf{Q}) = 
    \underset{ \mathbf{x} \in \{0,1\}^n }{\arg \min} \, \mathbf{x}^T \mathbf{Q}\mathbf{x}, 
\end{align}
where $\mathbf{x}$ is a binary vector for decision variable and $\mathbf{Q} \in \mathbb{R}^{n \times n}$ is a square matrix of constants associated with specific tasks which we will name as the $Q$-matrix in the following paper.


\subsection{Graph representation for CO problems}



\begin{figure}[t]
    \centering
    \includegraphics[width = \linewidth]{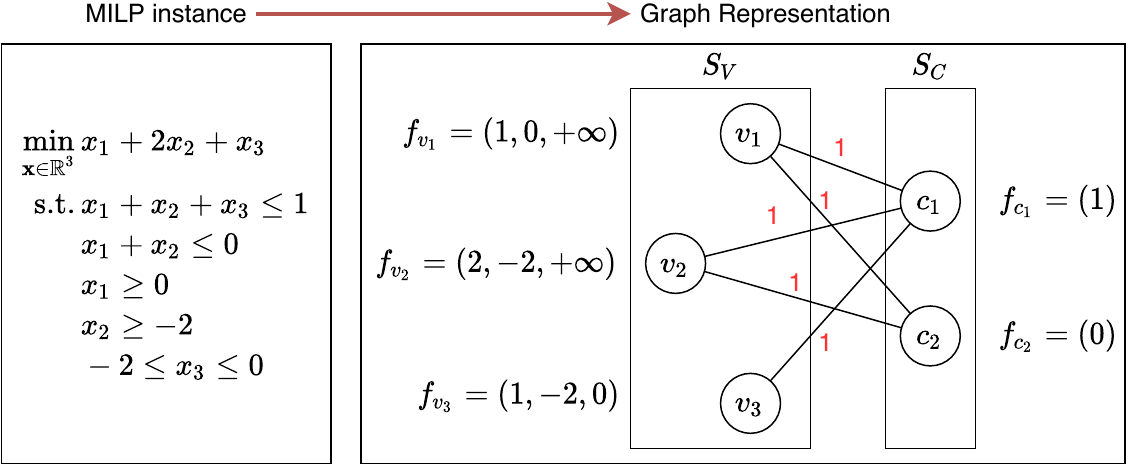}
    \caption{A simple example of weighted bipartite graph representation for an MILP instance. $S_V$ and $S_C$ denote the two node sets for variables and constraints, and $f_{v}/f_{c}$ denote the node feature.}
    \label{fig: MILP-example}
\end{figure}

In order to utilize GNNs to solve the above CO problems, we follow the previous literature \cite{Gasse2019ExactCO} to represent combinatorial optimization instances as a graph. Next, we introduce the details of representation of combinatorial optimization as graphs. 

\paragraph{Representing MILP.} Inspired by \cite{Gasse2019ExactCO}, we can model an MILP instance by a bipartite graph. Let's take equation \eqref{equation: MILP} as an example, the  bipartite graph is denoted as $\mathcal{G} = (S_V \cup S_C, \mathcal{E})$, where $S_V \cap S_C = \emptyset$. $S_V = \{v_1, v_2, \ldots, v_n\}$ denotes the variables $\mathbf{x} \in \mathbb{R}^n$ and $S_C = \{c_1, c_2, \ldots, c_m\}$ denotes the constraints. For better representing equation \eqref{equation: MILP} in graph $\mathcal{G}$, we add several handcraft features to the nodes and edges. The fixed information for node $v_i$ is a feature vector $f_{v_i} = (c_i, l_i, u_i, \tau_i) \in \mathbb{R} \times (\mathbb{R} \cup \{-\infty\}) \times (\mathbb{R} \cup \{+\infty\}) \times \{0,1\} $, where $\tau_i = 1$ if $\mathbf{x}_i \in \mathbb{Z}$, otherwise, $\tau_i = 0$. The feature for node $c_j$ is $b_j \in \mathbb{R}$. And the responding weight on edges $(v_i, c_j) \in \mathcal{E}$ is $\mathbf{A}_{ij}$. Such a weighted bipartite graph $\mathcal{G} = (S_V \cup S_C, \mathcal{E})$ can cover all the information in MILP given in equation \eqref{equation: MILP}. We provide a simple example in \Cref{fig: MILP-example} to illustrate the above graph representation method for MILP.

\begin{figure}[t]
    \centering
    \includegraphics[width = \linewidth]{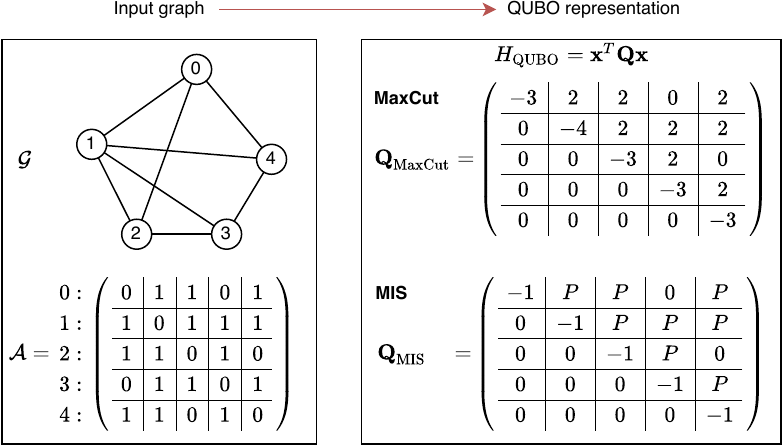}
    \caption{A simple example of QUBO formulation for a normal graph. $P$ is a fixed penalty parameter for the problems and $\mathbf{Q}$ is the $Q$-matrix for different tasks in equation \eqref{equation: QUBO}.}
    \label{fig: QUBO-example}
\end{figure}

\paragraph{Representing QUBO.} Inspired by \cite{Schuetz2021CombinatorialOW}, the form of QUBO can represent a large of canonical NP-hard problems including network combinatorial optimization (e.g. MaxCut, MIS, MVC) whose input is a graph. As illustrated in previous examples (see \Cref{example: MaxCut}, \Cref{example: MIS}), we can see numerous essential problems fall into the QUBO class naturally \cite{Glover2022QuantumBA}. In \Cref{fig: QUBO-example}, we provide a simple example of QUBO formulation associated with specific tasks (e.g. MaxCut and MIS).

\section{Framework}


\subsection{Graph Neural Architecture Search}

Graph neural architecture search (GraphNAS) is now recognized as an effective tool for generating new GNNs. Following the previous literature \cite{Liu2018DARTSDA,DBLP:conf/nips/WangGSYY22}, we model the GraphNAS as a bi-level optimization as follows, 

    \vspace{-0.5cm}
\begin{align}\label{equation: GNAS}
    \min_{\mathbf{a}} \quad &\mathcal{L}_{val}(\mathbf{w}^*(\mathbf{a}), \mathbf{a}) \\
    \text{s.t.} \quad &\mathbf{w}^*(\mathbf{a}) = \arg \min_{\mathbf{w}} \mathcal{L}_{train}(\mathbf{w}, \mathbf{a}), 
\end{align}
where $\mathbf{a}, \mathbf{w}$ denote the  architecture parameters and operation weights respectively. $\mathcal{L}$ denotes a supervised loss function, i.e. mean squared error, cross-entropy.

The key step is to design the search space $\{a\}$. Most GNNs conform to the messing passing between neighboring nodes \cite{kipf2017semi}, which can be described as the following iteration functions, 

    \vspace{-0.5cm}
\begin{align}\label{equation: GNN-MP}
    m_{v}^{l+1} &= f_{\theta}^{l} ( h_{v}^{l}, \{h_{u}^{l} | u \in \mathcal{N}_{v}\} )\\
    h_{v}^{l+1} &= \sigma^{l} ( g^{l}( h_{v}^{l},m_{v}^{l+1}  ) ), 
\end{align}
where $f_{\theta}^{l}, \sigma^{l}, g^{l}$ denote parametric functions, i.e., neighborhood aggregation function, activation function (e.g. sigmoid, ReLU), and combination function (e.g. summation, mean), at the $l$-layer messaging passing on graphs. $\mathcal{N}_{v}$ denotes the neighborhood for node $v$ and $h_{v}^{l}$ denotes the hidden embedding for $v$. Such message passing  in equation \eqref{equation: GNN-MP} will repeat $L$ times ($l \in \{1, 2, \ldots, L\}$) until converge. For the combinatorial optimization task in this paper, the information will only pass across the entire graph.

Taking GCN for example,  the message-passing equation in  graph convolutional network \cite{kipf2017semi} is explicitly written as
    \begin{align*}
        h_{v}^{l+1} = \sigma \left( \mathbf{W}^l \sum_{u \in \mathcal{N}_{v}} \frac{h_{u}^{l}}{|\mathcal{N}_{v}|} + \mathbf{B}^l h_{v}^{l}  \right),
    \end{align*}
where $\mathbf{W}^l$ and $\mathbf{B}^l$ are learnable parameters in $l$-layer. Most GNN-based frameworks involve GCN as the default architecture, but our task is to find a better task-focused architecture, at least better than GCN.

\subsection{AutoGNP}

In recent years, there exist various graph neural architecture frameworks and they have obtained promising popularity in many applications like node classification and link prediction. Almost all the automatic systems in this domain ignore the NP-hard CO problems learning task.

In this part, we introduce the AutoGNP method, which a differentiable GraphNAS framework for solving NP-hard combinatorial optimization problems. For efficiency, we also leverage a bi-level optimization in AutoGNP following previous differentiable algorithms \cite{Liu2018DARTSDA,Zhao2021SearchTA} similar to equation \eqref{equation: GNAS} as follows:
\begin{align}\label{equation: bi-level}
    \min_{\mathbf{a}} \quad &\mathcal{M}_{val}(\mathbf{w}^*(\mathbf{a}), \mathbf{a}) \\
    \text{s.t.} \quad &\mathbf{w}^*(\mathbf{a}) = \arg \min_{\mathbf{w}} \mathcal{M}_{train}(\mathbf{w}, \mathbf{a}), 
\end{align}
where $\mathcal{M}$ denotes the metric to qualify a combinatorial optimization task in the training. For example, in QUBO modeled problem, $\mathcal{M}$ can be expressed as a Hamiltonian loss, i.e.,  
   $ \mathcal{M}(\mathbf{x}) = \mathbf{x}^T \mathbf{Q} \mathbf{x}$. 

\begin{table*}[t]
\small
\caption{Search space $\mathcal{O}$ for architecture component and the operations of GNN we used in AutoGNP, which are commonly used as functions and hyperparameters in the GNN community.}
\label{table: operations}
\vskip 0.05in
\centering
\begin{center}
\begin{tabular}{|l|l|p{6cm}|}
\hline
\multicolumn{1}{|c|}{\bf Operation type $\mathcal{O}$} 
&\multicolumn{1}{|c|}{\bf Descriptions} 
&\multicolumn{1}{c|}{\bf Candidate operations}
\\ \hline 
$\mathcal{O}_a$         
& Attention function
&\texttt{CNN, GCN, GAT, GAT-SYM, GAT-COS, GAT-LINEAR, GAT-GEN-LINEAR, GraphSAGE-MEAN, GraphSAGE-MAX, GIN} \\ \hline
$\mathcal{O}_{agg}$             
& Aggregation function
&\texttt{SUM, MAX-POOLING, MEAN-POOLING, CONCAT, RNN, LSTM} 
\\ \hline
$\mathcal{O}_s$             
& Skip connection
&\texttt{IDENTITY, ZERO, STACK, SKIP-SUM, SKIP-CAT}
\\ \hline
$\mathcal{O}_c$             
& Combine function
&\texttt{AVG, CONCAT}
\\ \hline
\end{tabular}
\end{center}
\end{table*}

{\bf Design for $\mathcal{M}$.} The difference between AutoGNP and existing GraphNAS frameworks is the metric $\mathcal{M}$ to measure GNN models. In our CO problems, we usually do not have predefined loss functions to guide the training process. We should note that our objective function may achieve an unexpected minimum and we can involve two optimization tricks for better training as follows:
\begin{itemize}
    \item \textit{Simulated annealing for optimization}. Details see the above equation \eqref{noise}, we would leverage simulated annealing for obtaining a global minimum.
    \item \textit{Strict early stopping policy}. Different from default adaptive early stopping, we use a more strict policy for efficiency and set the stopping patience and threshold as constants in order to avoid failing to local minima.
    \end{itemize}

{\bf Differentiable search algorithm.} Compared with other reinforcement learning based frameworks \cite{Gao2019GraphNASGN,Jiang2020GraphNN}, we here consider a more efficient differentiable algorithms \cite{Zhao2021SearchTA,Wang2021AutoGELAA} for the search algorithm in equation \eqref{equation: GNAS}. In \Cref{table: properties}, we classify existing similar GraphNAS from several properties, such as end-to-end, efficient, differentiable, search algorithm, and whether supports CO problems. According to the optimization methods for our framework, we involve a noise-based trick to avoid failure into a local minimum. For example, when the gradient and function value are both close to zero, we prefer to add some Gaussian noise $\mathbf{ \epsilon } \sim \mathcal{N}(0, I)$ to the function value in order to jump outside the local minimum proactively such as
\begin{align}\label{noise}
    \mathcal{M}(\mathbf{x}_t) &= ( \mathbf{ x }_t + \mathbf{ \epsilon } )^T \mathbf{Q} ( \mathbf{ x }_t + \mathbf{ \epsilon } ), \\  \text{ when } & \|\mathbf{ x }_t\|\leq \delta, \|\partial \mathcal{M}(\mathbf{x}_t) \|\leq \delta, \nonumber
\end{align}
where $\delta$ is a real number close to zero. This trick is essential for the AutoGNP search framework because most combinatorial optimization has a large number of solutions that have similar qualified performance to the best. Also, they may have an obvious local solution that satisfies the $\|\partial \mathcal{M}(\mathbf{x}_t) \| = 0$ but it is not a correct solution to the problem. Take MaxCut as an example, $\mathbf{ x } = \mathbf{ 0 }$ is a solution such that Hamiltonian loss $H_{\text{MaxCut}} = 0$ in equation \eqref{equation:MaxCut}.

Here, we give a comprehensive analysis of AutoGNP from the view of optimization. Following the gradient-based approximation in \cite{huan2021search}, we update the variable $\mathbf{a}$ in $\mathcal{M}_{val}(\mathbf{w}^*(\mathbf{a}), \mathbf{a})$ by approximating the gradient $\nabla_{\mathbf{a}}\mathcal{M}_{val}(\mathbf{w}^*(\mathbf{a}), \mathbf{a})$ as follows:
\begin{align} 
    \nabla_{\mathbf{a}}\mathcal{M}_{val}(\mathbf{w} - \xi \nabla_{\mathbf{w}}\mathcal{M}_{train}(\mathbf{w}, \mathbf{a}) , 
    \mathbf{a}),
\end{align}
where $\mathbf{w}$ is the current operation weight and $\xi$ is the learning rate of the inner optimization. For the training complexity, we are similar to the SANE and we can achieve almost the same efficiency in our focused tasks. Compared with the existing GraphNAS framework, we provide \Cref{table: properties} to show superior efficiency and NP-hard tasks.

\begin{table*}[t]
\caption{ Numerical results for MaxCut on Gset instances. Our results can exceed PI-GNN and are close to BLS (best solutions for Gset), demonstrating the state-of-the-art capability to use GNNs for Gset. The bold letters denote the best solutions of BLS and the underline letters remark the best solutions except the BLS.}
    \label{tab:MaxCut}
    \vskip 0.05in
    \centering
    \begin{tabular}{|c|cc|ccccc|c|}
\hline \text { graph } & \text { nodes } & \text { edges } & \text { BLS } & \text { DSDP } & \text { KHLWG } & \text { RUN-CSP } & \text { PI-GNN } & \text { AutoGNP } \\
\hline \hline \text { G14 } & 800 & 4694 & \textbf{3064} & 2922 & 3061 & 2943 & 3026 &  \underline{3062}\\
\text { G15 } & 800 & 4661 & \textbf{3050} & 2938 & \underline{3050} & 2928 & 2990 &  3040\\
\text { G22 } & 2000 & 19990 & \textbf{13359} & 12960 & \underline{13359} & 13028 & 13181 &  13333 \\
\text { G49 } & 3000 & 6000 & \textbf{6000} &  \underline{6000} &  \underline{6000} &  \underline{6000} & 5918 &  \underline{6000} \\
\text { G50 } & 3000 & 6000 & \textbf{5880} &  \underline{5880} &  \underline{5880} &  \underline{5880} & 5820 &   \underline{5880}\\
\text { G55 } & 5000 & 12468 & \textbf{10294} & 9960 &  \underline{10236} & 10116 & 10138 & 10162 \\
\text { G70 } & 10000 & 9999 & \textbf{9541} & 9456 & 9458 & - & 9421 &  \underline{9499} \\
\hline
    \end{tabular}
\end{table*}


{\bf Search space for GNN architecture.} To design a better search space $\mathcal{O}$ for AutoGNP, we collect almost all the state-of-the-art candidate operators for graph neural networks, especially for the task of combinatorial optimization. We notice that most similar works only use the ordinary message-passing GNNs \cite{Schuetz2021CombinatorialOW,Gasse2019ExactCO}. However, we consider a diverse range among the architectures for graph layers. In \Cref{table: operations}, we summarize candidate choices for those search spaces and part of the operators followed by \cite{Zhao2020SimplifyingAS,Gao2019GraphNASGN}. Furthermore, we design a two-hop operator for the NP-hard problem which can be regarded as a generalized neighbor for nodes when considering the aggregation operators. Similar to the aggregation function in equation \eqref{equation: GNN-MP}, the two-hop message passing functions to combinatorial optimization are designed as follows:
\begin{align}\label{2-hop}
    m_{v}^{l+1} &= f_{\theta}^{l} ( h_{v}^{l}, \{h_{u}^{l} | u \in \mathcal{N}^{(2)}_{v}\} ),
\end{align}
where $\mathcal{N}^{(2)}_{v}$ denotes the two-hop neighbors to node $v$ which can also be written as a node set as follows:
\begin{align*}
    \mathcal{N}^{(2)}_{v} = \{ v' \mid  \|\mathcal{N}_{v} \cap \mathcal{N}_{v'}\| > 0 \}.
\end{align*}

{\bf Search space for hyperparameters.} In \Cref{table: HyperparameterRange}, it provides the range of hyper parameters. This table includes the range for common learning parameters such as learning rate, optimizer, the number of training epochs, and batch size. And ``2-hop'' means whether it contains the special operator mentioned in equation \eqref{2-hop} for aggregations in GNNs. In such a large range of hyperparameters, we can train a better GNN when given the architecture.

\section{Experiments}\label{section:experiments}

\begin{table*}[t]
\small
\caption{The range of hyper parameters.}
\label{table: HyperparameterRange}
\vskip 0.05in
\centering
\begin{tabular}{|l|p{9cm}|}
\hline
\multicolumn{1}{|l|}{\bf Hyper parameter} 
&\multicolumn{1}{c|}{\bf Range/Choices}
\\ \hline 
Initialization & Xavier, Kaiming, Uniform \\ \hline
Learning rate & $0.0001 \sim 0.1000$\\ \hline
Batch normalization & True, False \\ \hline
Optimizer & ADAM, ADAMW \\ \hline
\# Training epoch & $2 \sim 300$ \\ \hline
Batch size & 8, 16, 32 \\ \hline
Stopping patience & $3 \sim 7$\\ \hline
2-hop & True, False \\ \hline
Activation function
&\texttt{SIGMOD, THAN, RELU, LINEAR, LEAKY-RELU, ELU}
\\ \hline
\# Attention head
&\texttt{1, 2, 4, 6, 8, 16, 32}
\\ \hline
\# GNN layers
&
\texttt{2, 3, 4, 5, 6}
\\ \hline
Hidden dimension
&\texttt{2, 4, 8, 16, 32, 64, 128, 256, 512}
\\ \hline
\end{tabular}
\end{table*}


\subsection{Benchmark datasets}

We evaluate AutoGNP on several NP-hard datasets following previous works \cite{Gasse2019ExactCO,Glover2022QuantumBA} and we summarize the graph representations for each task in \Cref{table: RepresentationTasks}.

\begin{itemize}
    \item Set covering \cite{strm1965OptimalCO}. We use the training and testing instances for set covering with 1,000 columns.
    \item Combinatorial auction \cite{LeytonBrown2000TowardsAU}. We use the  combinatorial auction instances which are generated in \cite{LeytonBrown2000TowardsAU}
    \item Facility location \cite{Cornujols1991ACO}. We use the training and testing instances for set covering with 100 facilities.
    \item MaxCut. We have conducted supplementary experiments on Max-Cut benchmark instances and their work on random d-regular graphs. These experiments were performed using the publicly available Gset dataset\footnote{{https://web.stanford.edu/~yyye/yyye/Gset/}. 
    }. 
    \item MIS \cite{Cir2018DecisionDF}. We conduct our experiments on E-R random graphs followed by \cite{Cir2018DecisionDF}.
\end{itemize}

\subsection{Baselines}

We compare against several classical machine learning baselines as follows:
\begin{itemize}
    \item TREES \cite{alvarez2017machine}: a learning-to-score approach based on ExtraTrees models \cite{geurts2006extremely}.
    \item SVMRANK \cite{Khalil2016LearningTB}: a learning-to-rank approach based on SVMrank \cite{joachims2002optimizing}.
    \item LMART \cite{hansknecht2018cuts}: a learning-to-rank approach based on LambdaMART\cite{burges2010ranknet}.
    \item GCNN \cite{gasse2019exact}: graph convolutional neural network based branching policies that exploit the inherent bipartite graph representation of MILP problems.
\end{itemize}
The details of the baselines follow \cite{gasse2019exact} including the original features proposed by \cite{Khalil2016LearningTB}. Following the previous paper \cite{Schuetz2021CombinatorialOW}, we include some baseline methods for MaxCut including breakout local search (BLS) \cite{benlic2013breakout}, SDP solver (DSDP) \cite{choi2000solving}, tabu search metaheuristic (KHLWG) \cite{kochenberger2013solving}, recurrent GNN architecture (RUN-CSP) \cite{toenshoff2019run}, and physics-inspired GNNs (PI-GNN) \cite{Schuetz2021CombinatorialOW}. Note that BLS obtains the best solutions for Gset.

\begin{table}[b]
\caption{Representation of the tasks using MILP and QUBO.}
\label{table: RepresentationTasks}
\centering
\begin{tabular}{|l|p{4cm}|}
\hline
\multicolumn{1}{|l|}{\bf Representations} 
&\multicolumn{1}{c|}{\bf Tasks}
\\ \hline 
MILP & Set covering, Combinatorial auction, Facility location \\ \hline
QUBO & MaxCut, MIS
\\ \hline
\end{tabular}
\end{table}

\begin{table*}[t]
\small
    \caption{ {Performance (accuracy) on Set Covering and Combinatorial Auction. The metric is accuracy@$k$ (\%) ($k = 1,5,10$).} } \label{table: Performance1}
    \vskip 0.05in
    \centering
         \begin{tabular}{lcccccc} 
        \toprule
        & \multicolumn{3}{c}{\textbf{ Set Covering }}
        & \multicolumn{3}{c}{\textbf{  Combinatorial Auction }}
        \\
        \cmidrule(lr){2-4}
        \cmidrule(lr){5-7}
        model & acc@1 & acc@5 & acc@10 
        & acc@1 & acc@5 & acc@10\\
        \midrule
        TREES & 51.8 $\pm$ 0.3 &  80.5 $\pm$ 0.1 &  91.4 $\pm$ 0.2
            &  52.9 $\pm$ 0.3 &  84.3 $\pm$ 0.1 &  94.1 $\pm$ 0.1
            \\
        SVMRANK & 57.6 $\pm$ 0.2 & 84.7 $\pm$ 0.1 & 94.0 $\pm$ 0.1
            & 57.2 $\pm$ 0.2 & 86.9 $\pm$ 0.2 & 95.4 $\pm$ 0.1 \\
        LMART  & 57.4 $\pm$ 0.2 & 84.5 $\pm$ 0.1 & 93.8 $\pm$ 0.1
             & 57.3 $\pm$ 0.3 & 86.9 $\pm$ 0.2 & 95.3 $\pm$ 0.1
            \\
        GCNN  & 65.5 $\pm$ 0.1 & {92.4} $\pm$ 0.1 & 98.2 $\pm$ 0.0
         & 61.6 $\pm$ 0.1 & 91.0 $\pm$ 0.1 & \textbf{97.8} $\pm$ 0.1 
         \\
        \midrule
        AutoGNP 
        & \textbf{67.0} $\pm$ 0.1 & \textbf{92.6} $\pm$ 0.1 & \textbf{98.8} $\pm$ 0.0
        & \textbf{62.0} $\pm$ 0.2 & \textbf{91.6} $\pm$ 0.2 & \textbf{97.8} $\pm$ 0.1
        \\
        \bottomrule
         \end{tabular}
\end{table*}

\subsection{Experimental setup}

This section introduces the experimental setup for the comparative experiment, including the training details and the evaluation methods for all the NP-hard problems. Our implementation is based on the previous implementations of AutoGEL\footnote{https://github.com/zwangeo/AutoGEL.}, PI-GNN\footnote{https://github.com/amazon-science/co-with-gnns-example.} and GCNN
, including the basic training details and metrics. The search space for architecture and hyper parameters are in \Cref{table: operations} and \Cref{table: HyperparameterRange}. The best hyper parameters are found after finding the best GNN architecture for our tasks. We almost include the state-of-the-art graph operations in it. The learning framework follows \cite{Schuetz2021CombinatorialOW} and \cite{gasse2019exact}.

\subsection{Results and analysis}

\Cref{tab:MaxCut} provides the results of AutoGNP for MaxCut on the Gset datasets. It can be seen that the results can almost achieve the best solution in this domain and certainly outperform the PI-GNN which is only based on GCN baselines. We note that on the G49 and G50 datasets, AutoGNP can achieve the best solution as BLS. \Cref{table: Performance1} and \Cref{table: Performance2} provide the results for set covering, combinatorial auction, facility location, and maximum independent set. We note that all results are conducted over five seeds.

\begin{table*}[t]
\small
    \caption{ {Performance (accuracy) on  Maximum Independent Set and Facility Location. The metric is accuracy@$k$ (\%) ($k = 1,5,10$).} } \label{table: Performance2}
    \vskip 0.05in
    \centering
         \begin{tabular}{lcccccc} 
        \toprule
        & \multicolumn{3}{c}{\textbf{  Maximum Independent Set }}
        & \multicolumn{3}{c}{\textbf{  Facility Location }}
        \\
        \cmidrule(lr){2-4}
        \cmidrule(lr){5-7}
        model & acc@1 & acc@5 & acc@10 
        & acc@1 & acc@5 & acc@10 
        \\
        \midrule
        TREES & 30.9 $\pm$ 0.4 &  47.4 $\pm$ 0.3 &  54.6 $\pm$ 0.3 
        & 63.0 $\pm$ 0.4 &  97.3 $\pm$ 0.1 &  \textbf{99.9} $\pm$ 0.0\\
        SVMRANK & 48.0 $\pm$ 0.6 & 69.3 $\pm$ 0.2 & 78.1 $\pm$ 0.2 & 67.8 $\pm$ 0.1 & 98.1 $\pm$ 0.1 &  \textbf{99.9} $\pm$ 0.0\\
        LMART  & 48.9  $\pm$ 0.3 & 68.9 $\pm$ 0.4 & 77.0 $\pm$ 0.5
        & 68.0 $\pm$ 0.2 & 98.0 $\pm$ 0.0 & \textbf{99.9} $\pm$ 0.0\\
        GCNN & 56.5 $\pm$ 0.2 & 80.8 $\pm$ 0.3 & 89.0 $\pm$ 0.1
        & 71.2 $\pm$ 0.2 & 98.6 $\pm$ 0.1 & \textbf{99.9} $\pm$ 0.0\\
        \midrule
        AutoGNP & \textbf{58.1} $\pm$ 0.4 & \textbf{84.0} $\pm$ 0.3 & \textbf{90.1} $\pm$ 0.2
        & \textbf{71.9} $\pm$ 0.3 & \textbf{98.8} $\pm$ 0.3 & \textbf{99.9} $\pm$ 0.0\\
        \bottomrule
         \end{tabular}
\end{table*}

Here, we provide an overall analysis of our results and the superiority of AutoGNP. In almost all combinatorial optimization tasks, we can achieve the level of expert solvers. Also, we can use AutoGNP to amplify the capability of GNNs in solving NP-hard CO problems and the results show that our framework can find better architectures than PI-GNN and GCN given in baselines.

\section{Conclusions}

In this work, we focus on developing automated graph neural networks for solving NP-hard combinatorial optimization problems. We propose AutoGNP, which is a data-driven automatic architecture search framework of graph neural networks to solve two specific combinatorial optimization problems, i.e., mixed integer linear programming and quadratic unconstrained binary optimization. We leverage the powerful representation capability of GNNs to represent combinatorial optimization problems as graphs and design efficient training tricks for the given CO problems. AutoGNP provides a novel graph neural architecture search framework for solving NP-hard CO problems, effectively reducing the search time cost required to generate new graph neural architectures.  
Empirical results show that  AutoGNP can generate better GNN architectures for the given combinatorial optimization problems. Though there may exist challenges of distorted landscape and biased optimization in our training procedure which inherits the limitation from the differentiable architecture search algorithm DARTS \cite{Liu2018DARTSDA,huan2021search,wang2022zarts}, we leave this as the future work.

\bibliography{aaai24}

\begin{thebibliography}{49}
\providecommand{\natexlab}[1]{#1}

\bibitem[{Alvarez, Louveaux, and Wehenkel(2017)}]{alvarez2017machine}
Alvarez, A.~M.; Louveaux, Q.; and Wehenkel, L. 2017.
\newblock A machine learning-based approximation of strong branching.
\newblock \emph{INFORMS Journal on Computing}, 29(1): 185--195.

\bibitem[{{\AA}str{\"o}m(1965)}]{strm1965OptimalCO}
{\AA}str{\"o}m, K.~J. 1965.
\newblock Optimal control of Markov processes with incomplete state information.
\newblock \emph{Journal of Mathematical Analysis and Applications}, 10: 174--205.

\bibitem[{Bello et~al.(2016)Bello, Pham, Le, Norouzi, and Bengio}]{bello2016neural}
Bello, I.; Pham, H.; Le, Q.~V.; Norouzi, M.; and Bengio, S. 2016.
\newblock Neural combinatorial optimization with reinforcement learning.
\newblock \emph{arXiv preprint arXiv:1611.09940}.

\bibitem[{Bengio, Lodi, and Prouvost(2018)}]{Bengio2018MachineLF}
Bengio, Y.; Lodi, A.; and Prouvost, A. 2018.
\newblock Machine Learning for Combinatorial Optimization: a Methodological Tour d'Horizon.
\newblock \emph{Eur. J. Oper. Res.}, 290: 405--421.

\bibitem[{Benlic and Hao(2013)}]{benlic2013breakout}
Benlic, U.; and Hao, J.-K. 2013.
\newblock Breakout local search for the max-cutproblem.
\newblock \emph{Engineering Applications of Artificial Intelligence}, 26(3): 1162--1173.

\bibitem[{Berthold, Francobaldi, and Hendel(2022)}]{Berthold2022LearningTU}
Berthold, T.; Francobaldi, M.; and Hendel, G. 2022.
\newblock Learning to Use Local Cuts.

\bibitem[{Burges(2010)}]{burges2010ranknet}
Burges, C.~J. 2010.
\newblock From ranknet to lambdarank to lambdamart: An overview.
\newblock \emph{Learning}, 11(23-581): 81.

\bibitem[{Cappart et~al.(2023)Cappart, Ch{\'e}telat, Khalil, Lodi, Morris, and Velickovic}]{cappart2023combinatorial}
Cappart, Q.; Ch{\'e}telat, D.; Khalil, E.~B.; Lodi, A.; Morris, C.; and Velickovic, P. 2023.
\newblock Combinatorial optimization and reasoning with graph neural networks.
\newblock \emph{J. Mach. Learn. Res.}, 24: 130--1.

\bibitem[{Choi and Ye(2000)}]{choi2000solving}
Choi, C.; and Ye, Y. 2000.
\newblock Solving sparse semidefinite programs using the dual scaling algorithm with an iterative solver.
\newblock \emph{Manuscript, Department of Management Sciences, University of Iowa, Iowa City, IA}, 52242.

\bibitem[{Cir{\'e}(2018)}]{Cir2018DecisionDF}
Cir{\'e}, A.~A. 2018.
\newblock Decision diagrams for optimization.
\newblock \emph{Constraints}, 20: 494--495.

\bibitem[{Cornu{\'e}jols, Sridharan, and Thizy(1991)}]{Cornujols1991ACO}
Cornu{\'e}jols, G.; Sridharan, R.; and Thizy, J.-M. 1991.
\newblock A comparison of heuristics and relaxations for the capacitated plant location problem.
\newblock \emph{European Journal of Operational Research}, 50: 280--297.

\bibitem[{Ding et~al.(2021)Ding, Yao, Zhao, and Zhang}]{ding2021diffmg}
Ding, Y.; Yao, Q.; Zhao, H.; and Zhang, T. 2021.
\newblock Diffmg: Differentiable meta graph search for heterogeneous graph neural networks.
\newblock In \emph{Proceedings of the 27th ACM SIGKDD Conference on Knowledge Discovery \& Data Mining}, 279--288.

\bibitem[{Gao et~al.(2020)Gao, Yang, Zhang, Zhou, and Hu}]{Gao2019GraphNASGN}
Gao, Y.; Yang, H.; Zhang, P.; Zhou, C.; and Hu, Y. 2020.
\newblock GraphNAS: Graph Neural Architecture Search with Reinforcement Learning.

\bibitem[{Gasse et~al.(2019{\natexlab{a}})Gasse, Ch{\'e}telat, Ferroni, Charlin, and Lodi}]{gasse2019exact}
Gasse, M.; Ch{\'e}telat, D.; Ferroni, N.; Charlin, L.; and Lodi, A. 2019{\natexlab{a}}.
\newblock Exact combinatorial optimization with graph convolutional neural networks.
\newblock \emph{Advances in neural information processing systems}, 32.

\bibitem[{Gasse et~al.(2019{\natexlab{b}})Gasse, Ch{\'e}telat, Ferroni, Charlin, and Lodi}]{Gasse2019ExactCO}
Gasse, M.; Ch{\'e}telat, D.; Ferroni, N.; Charlin, L.; and Lodi, A. 2019{\natexlab{b}}.
\newblock Exact Combinatorial Optimization with Graph Convolutional Neural Networks.
\newblock In \emph{Neural Information Processing Systems}.

\bibitem[{Geurts, Ernst, and Wehenkel(2006)}]{geurts2006extremely}
Geurts, P.; Ernst, D.; and Wehenkel, L. 2006.
\newblock Extremely randomized trees.
\newblock \emph{Machine learning}, 63: 3--42.

\bibitem[{Glover et~al.(2022)Glover, Kochenberger, Hennig, and Du}]{Glover2022QuantumBA}
Glover, F.~W.; Kochenberger, G.~A.; Hennig, R.; and Du, Y. 2022.
\newblock Quantum bridge analytics I: a tutorial on formulating and using QUBO models.
\newblock \emph{Annals of Operations Research}, 314: 141 -- 183.

\bibitem[{Hammouri et~al.(2018)Hammouri, Samra, Al-Betar, Khalil, Alasmer, and Kanan}]{hammouri2018dragonfly}
Hammouri, A.~I.; Samra, E. T.~A.; Al-Betar, M.~A.; Khalil, R.~M.; Alasmer, Z.; and Kanan, M. 2018.
\newblock A dragonfly algorithm for solving traveling salesman problem.
\newblock In \emph{2018 8th IEEE international conference on control system, computing and engineering (ICCSCE)}, 136--141. IEEE.

\bibitem[{Hansknecht, Joormann, and Stiller(2018)}]{hansknecht2018cuts}
Hansknecht, C.; Joormann, I.; and Stiller, S. 2018.
\newblock Cuts, primal heuristics, and learning to branch for the time-dependent traveling salesman problem.
\newblock \emph{arXiv preprint arXiv:1805.01415}.

\bibitem[{Huan, Quanming, and Weiwei(2021)}]{huan2021search}
Huan, Z.; Quanming, Y.; and Weiwei, T. 2021.
\newblock Search to aggregate neighborhood for graph neural network.
\newblock In \emph{2021 IEEE 37th International Conference on Data Engineering (ICDE)}, 552--563. IEEE.

\bibitem[{Jiang and Balaprakash(2020)}]{Jiang2020GraphNN}
Jiang, S.; and Balaprakash, P. 2020.
\newblock Graph Neural Network Architecture Search for Molecular Property Prediction.
\newblock \emph{2020 IEEE International Conference on Big Data (Big Data)}, 1346--1353.

\bibitem[{Joachims(2002)}]{joachims2002optimizing}
Joachims, T. 2002.
\newblock Optimizing search engines using clickthrough data.
\newblock In \emph{Proceedings of the eighth ACM SIGKDD international conference on Knowledge discovery and data mining}, 133--142.

\bibitem[{Khalil et~al.(2016)Khalil, Bodic, Song, Nemhauser, and Dilkina}]{Khalil2016LearningTB}
Khalil, E.~B.; Bodic, P.~L.; Song, L.; Nemhauser, G.~L.; and Dilkina, B.~N. 2016.
\newblock Learning to Branch in Mixed Integer Programming.
\newblock In \emph{AAAI Conference on Artificial Intelligence}.

\bibitem[{Kipf and Welling(2017)}]{kipf2017semi}
Kipf, T.~N.; and Welling, M. 2017.
\newblock Semi-Supervised Classification with Graph Convolutional Networks.
\newblock In \emph{International Conference on Learning Representations (ICLR)}.

\bibitem[{Kochenberger et~al.(2013)Kochenberger, Hao, L{\"u}, Wang, and Glover}]{kochenberger2013solving}
Kochenberger, G.~A.; Hao, J.-K.; L{\"u}, Z.; Wang, H.; and Glover, F. 2013.
\newblock Solving large scale max cut problems via tabu search.
\newblock \emph{Journal of Heuristics}, 19: 565--571.

\bibitem[{Lai et~al.(2020)Lai, Zha, Zhou, and Hu}]{lai2020policy}
Lai, K.-H.; Zha, D.; Zhou, K.; and Hu, X. 2020.
\newblock Policy-gnn: Aggregation optimization for graph neural networks.
\newblock In \emph{Proceedings of the 26th ACM SIGKDD International Conference on Knowledge Discovery \& Data Mining}, 461--471.

\bibitem[{Leyton-Brown, Pearson, and Shoham(2000)}]{LeytonBrown2000TowardsAU}
Leyton-Brown, K.; Pearson, M.; and Shoham, Y. 2000.
\newblock Towards a universal test suite for combinatorial auction algorithms.
\newblock In \emph{ACM Conference on Economics and Computation}.

\bibitem[{Liu, Simonyan, and Yang(2018)}]{Liu2018DARTSDA}
Liu, H.; Simonyan, K.; and Yang, Y. 2018.
\newblock DARTS: Differentiable Architecture Search.
\newblock \emph{ArXiv}, abs/1806.09055.

\bibitem[{Liu et~al.(2023)Liu, Zhou, Zhang, Zhang, Zhang, Li, and Chen}]{liu2021decision}
Liu, Y.; Zhou, C.; Zhang, P.; Zhang, S.; Zhang, X.; Li, Z.; and Chen, H. 2023.
\newblock Decision-focused Graph Neural Networks for Graph Learning and Optimization.
\newblock In \emph{2023 IEEE International Conference on Data Mining (ICDM)}. IEEE.

\bibitem[{Morrison et~al.(2016)Morrison, Jacobson, Sauppe, and Sewell}]{Morrison2016BranchandboundAA}
Morrison, D.~R.; Jacobson, S.~H.; Sauppe, J.~J.; and Sewell, E.~C. 2016.
\newblock Branch-and-bound algorithms: A survey of recent advances in searching, branching, and pruning.
\newblock \emph{Discret. Optim.}, 19: 79--102.

\bibitem[{Nazari et~al.(2018)Nazari, Oroojlooy, Snyder, and Tak{\'a}c}]{nazari2018reinforcement}
Nazari, M.; Oroojlooy, A.; Snyder, L.; and Tak{\'a}c, M. 2018.
\newblock Reinforcement learning for solving the vehicle routing problem.
\newblock \emph{Advances in neural information processing systems}, 31.

\bibitem[{Nazari, Tarzanagh, and Michailidis(2022)}]{nazari2022dadam}
Nazari, P.; Tarzanagh, D.~A.; and Michailidis, G. 2022.
\newblock Dadam: A consensus-based distributed adaptive gradient method for online optimization.
\newblock \emph{IEEE Transactions on Signal Processing}, 70: 6065--6079.

\bibitem[{Paulus and Krause(2023)}]{Paulus2023LearningTD}
Paulus, M.~B.; and Krause, A. 2023.
\newblock Learning To Dive In Branch And Bound.
\newblock \emph{ArXiv}, abs/2301.09943.

\bibitem[{Ross(2013)}]{Ross2013InteractiveLF}
Ross, S. 2013.
\newblock Interactive Learning for Sequential Decisions and Predictions.

\bibitem[{Schuetz, Brubaker, and Katzgraber(2021)}]{Schuetz2021CombinatorialOW}
Schuetz, M. J.~A.; Brubaker, J.~K.; and Katzgraber, H.~G. 2021.
\newblock Combinatorial optimization with physics-inspired graph neural networks.
\newblock \emph{Nature Machine Intelligence}, 4: 367 -- 377.

\bibitem[{Toenshoff et~al.(2019)Toenshoff, Ritzert, Wolf, and Grohe}]{toenshoff2019run}
Toenshoff, J.; Ritzert, M.; Wolf, H.; and Grohe, M. 2019.
\newblock RUN-CSP: unsupervised learning of message passing networks for binary constraint satisfaction problems.
\newblock \emph{CoRR, abs/1909.08387}.

\bibitem[{Verma, Pant, and Snasel(2021)}]{verma2021comprehensive}
Verma, S.; Pant, M.; and Snasel, V. 2021.
\newblock A comprehensive review on NSGA-II for multi-objective combinatorial optimization problems.
\newblock \emph{Ieee Access}, 9: 57757--57791.

\bibitem[{Vinyals, Fortunato, and Jaitly(2015)}]{vinyals2015pointer}
Vinyals, O.; Fortunato, M.; and Jaitly, N. 2015.
\newblock Pointer networks.
\newblock \emph{Advances in neural information processing systems}, 28.

\bibitem[{Wang and Tang(2021)}]{wang2021deep}
Wang, Q.; and Tang, C. 2021.
\newblock Deep reinforcement learning for transportation network combinatorial optimization: A survey.
\newblock \emph{Knowledge-Based Systems}, 233: 107526.

\bibitem[{Wang et~al.(2022{\natexlab{a}})Wang, Guo, Su, Yang, and Yan}]{wang2022zarts}
Wang, X.; Guo, W.; Su, J.; Yang, X.; and Yan, J. 2022{\natexlab{a}}.
\newblock Zarts: On zero-order optimization for neural architecture search.
\newblock \emph{Advances in Neural Information Processing Systems}, 35: 12868--12880.

\bibitem[{Wang et~al.(2022{\natexlab{b}})Wang, Guo, Su, Yang, and Yan}]{DBLP:conf/nips/WangGSYY22}
Wang, X.; Guo, W.; Su, J.; Yang, X.; and Yan, J. 2022{\natexlab{b}}.
\newblock {ZARTS:} On Zero-order Optimization for Neural Architecture Search.
\newblock In \emph{NeurIPS}.

\bibitem[{Wang et~al.(2023)Wang, Balasubramanian, Ma, and Razaviyayn}]{wang2023zeroth}
Wang, Z.; Balasubramanian, K.; Ma, S.; and Razaviyayn, M. 2023.
\newblock Zeroth-order algorithms for nonconvex--strongly-concave minimax problems with improved complexities.
\newblock \emph{Journal of Global Optimization}, 87(2): 709--740.

\bibitem[{Wang, Di, and Chen(2021)}]{Wang2021AutoGELAA}
Wang, Z.; Di, S.; and Chen, L. 2021.
\newblock AutoGEL: An Automated Graph Neural Network with Explicit Link Information.
\newblock \emph{ArXiv}, abs/2112.01064.

\bibitem[{Xu et~al.(2023)Xu, Zhang, Liu, Sun, Zhao, Yang, and Yu}]{Xu2023DoNT}
Xu, P.; Zhang, L.; Liu, X.; Sun, J.; Zhao, Y.; Yang, H.; and Yu, B. 2023.
\newblock Do Not Train It: A Linear Neural Architecture Search of Graph Neural Networks.
\newblock \emph{ArXiv}, abs/2305.14065.

\bibitem[{You, Ying, and Leskovec(2020)}]{you2020design}
You, J.; Ying, Z.; and Leskovec, J. 2020.
\newblock Design space for graph neural networks.
\newblock \emph{Advances in Neural Information Processing Systems}, 33: 17009--17021.

\bibitem[{Zhang, Wang, and Zhu(2021)}]{Zhang2021AutomatedML}
Zhang, Z.; Wang, X.; and Zhu, W. 2021.
\newblock Automated Machine Learning on Graphs: A Survey.
\newblock \emph{ArXiv}, abs/2103.00742.

\bibitem[{Zhao, Wei, and Yao(2020)}]{Zhao2020SimplifyingAS}
Zhao, H.; Wei, L.; and Yao, Q. 2020.
\newblock Simplifying Architecture Search for Graph Neural Network.
\newblock \emph{ArXiv}, abs/2008.11652.

\bibitem[{Zhao, Yao, and Tu(2021)}]{Zhao2021SearchTA}
Zhao, H.; Yao, Q.; and Tu, W.-W. 2021.
\newblock Search to aggregate neighborhood for graph neural network.
\newblock \emph{2021 IEEE 37th International Conference on Data Engineering (ICDE)}, 552--563.

\bibitem[{Zhou et~al.(2022)Zhou, Huang, Song, Chen, and Hu}]{zhou2022auto}
Zhou, K.; Huang, X.; Song, Q.; Chen, R.; and Hu, X. 2022.
\newblock Auto-gnn: Neural architecture search of graph neural networks.
\newblock \emph{Frontiers in big Data}, 5: 1029307.

\end{thebibliography}

\appendix
\section*{Appendix}

\section{Notation table}

\begin{table}[h]
\small
    \caption{ \small{Notation table.} } \label{table: notation}
    \vskip 0.05in
    \centering
         \begin{tabular}{lll} 
        \toprule
        Domain 
        & Name & Description \\
        \midrule
        \multirow{4}{*}{{ Graph-based model }}
        & $\mathcal{G}$ & Input graph \\
        & $\mathcal{E}$ & Edge set \\
        & $\mathcal{V}$ & Node set \\
        & $\mathcal{A}$ & Adjacency matrix \\
        \midrule
        \multirow{3}{*}{{ NAS-based model }}
        & $\mathbf{a}$ & Architecture parameters \\
        & $\mathbf{w}$ & Operation weights \\
        & $\mathcal{O}$ & Search space \\
        \bottomrule
         \end{tabular}
\end{table}

\section{Examples of QUBO: MaxCut and MIS}

\begin{example}[Maximum cut (MaxCut)]\label{example: MaxCut}
    Given a graph $\mathcal{G} = (\mathcal{V}, \mathcal{E})$, a maximum cut, denoted as $\mathcal{V}^*$, is a cut whose size equals or surpasses that of any other cuts. That is, it is a partition of the set $\mathcal{V}$ into two sets $\mathcal{V}^*$ and $\mathcal{V}^{*C} = \mathcal{V}/{\mathcal{V}^*}$, such that the number of edges between $\mathcal{V}^*$ and $\mathcal{V}^{*C}$ exceeds that of any other partitions. The task of finding such a partition is referred to as the MaxCut problem. The Hamiltonian is as follows:

    \vspace{-0.5cm}
    \begin{align}\label{equation:MaxCut}
        H_{\text{MaxCut}} = \sum_{i<j} A_{ij}(2x_ix_j-x_i-x_j).
    \end{align}
\end{example}

\begin{example}[Maximum independent set (MIS)]\label{example: MIS}
    Given an undirected graph $\mathcal{G} = (\mathcal{V}, \mathcal{E})$, an independent set is defined such that any two points in the set are not connected by edges. The task of identifying the largest independent set in the graph $\mathcal{G}$ is referred to as the maximum independent set problem. The Hamiltonian is as follows, 
    
    \vspace{-0.5cm}
    \begin{align*}
        H_{\text{MIS}} = -\sum_{i \in \mathcal{V}} x_i + P \sum_{(i,j)\in \mathcal{E}} x_ix_j,
    \end{align*}
    where $P>0$ is a fixed penalty parameter.
\end{example}




\end{document}